\documentclass[letterpaper, 10 pt, conference]{ieeeconf}
\usepackage{xcolor}
\usepackage{booktabs}
\usepackage{makecell}
\usepackage{bm}
\usepackage{amsmath}
\usepackage{verbatim}
\usepackage{graphicx}
\usepackage{url}
\usepackage{amssymb}
\usepackage{tabularx}
\usepackage{multirow}
\usepackage{threeparttable}
\IEEEoverridecommandlockouts                              % This command is only needed if 
\title{\LARGE \bf
WS-DETR: Robust Water Surface Object Detection through Vision-Radar Fusion with Detection Transformer
}

\author{Huilin Yin$^{1}$, Pengyu Wang$^{1}$, Senmao Li$^{1}$, Jun Yan$^{1}$, and Daniel Watzenig$^{2}$% <-this % stops a space
\thanks{$^{1}$Huilin Yin, Pengyu Wang, Senmao Li, and Jun Yan are with the College of Electronics and Information Engineering, Tongji University, Shanghai 201804, China. Emails: \{yinhuilin, perry0817, lisenmao, yanjun\}@tongji.edu.cn}%
\thanks{$^{2}$Daniel Watzenig is with the Graz University of Technology and the Virtual Vehicle Research, Graz 8010, Austria. Emails: daniel.watzenig@tugraz.at}
}
\begin{document}

\maketitle
\thispagestyle{empty}
\pagestyle{empty}

\begin{abstract}
Robust object detection for Unmanned Surface Vehicles (USVs) in complex water environments is essential for reliable navigation and operation. Specifically, water surface object detection faces challenges from blurred edges and diverse object scales. Although vision-radar fusion offers a feasible solution, existing approaches suffer from cross-modal feature conflicts, which negatively affect model robustness. To address this problem, we propose a robust vision-radar fusion model WS-DETR. In particular, we first introduce a Multi-Scale Edge Information Integration (MSEII) module to enhance edge perception and a Hierarchical Feature Aggregator (HiFA) to boost multi-scale object detection in the encoder. Then, we adopt self-moving point representations for continuous convolution and residual connection to efficiently extract irregular features under the scenarios of irregular point cloud data. To further mitigate cross-modal conflicts, an Adaptive Feature Interactive Fusion (AFIF) module is introduced to integrate visual and radar features through geometric alignment and semantic fusion. Extensive experiments on the WaterScenes dataset demonstrate that WS-DETR achieves state-of-the-art (SOTA) performance, maintaining its superiority even under adverse weather and lighting conditions.
\begin{comment}
The code is available at \url{https://github.com/malice960/WS-DETR}.
\end{comment}

\end{abstract}

%%%%%%%%%%%%%%%%%%%%%%%%%%%%%%%%%%%%%%%%%%%%%%%%%%%%%%%%%%%%%%%%%%%%%%%%%%%%%%%%
\section{INTRODUCTION}

Unmanned Surface Vehicles (USVs) play a critical role in intelligent transportation systems~\cite{robot}, aquatic rescue operations~\cite{rescue}, and monitoring of the water environment~\cite{waterquality}. In these operational scenarios, advanced object detection systems are essential for avoiding obstacles, ensuring navigation safety, and improving operational efficiency. Compared to Unmanned Ground Vehicles (UGVs), Unmanned Surface Vehicles (USVs) face unique challenges~\cite{Mask-VRDet}, including blurred edges, significant variations in object scales on water surfaces, and highly dynamic operating environments. These conditions are often characterized by frequent adverse weather and lighting events, which can degrade sensor performance and impose more stringent requirements on the model’s environmental perception capabilities.

Although vision-based object detection systems provide high-resolution image and rich semantic information, their performance significantly degrades under adverse lighting or weather conditions. In contrast, mmWave radar demonstrates unique advantages through its strong signal penetration and Doppler velocity measurement capabilities, maintaining operational robustness in rain, fog, and other adverse environments~\cite{mmwave}. Recent advancements in semiconductor technology have enabled 4D-mmWave radar systems~\cite{reviewtiv}, which overcome the resolution limitations of traditional radar by delivering point cloud data that includes elevation information. The fusion of vision and 4D-mmWave radar presents a cost-effective solution for water surface perception, offering new possibilities for USV navigation systems.

\begin{figure}[!t]
    \centering
    \includegraphics[width=\columnwidth]{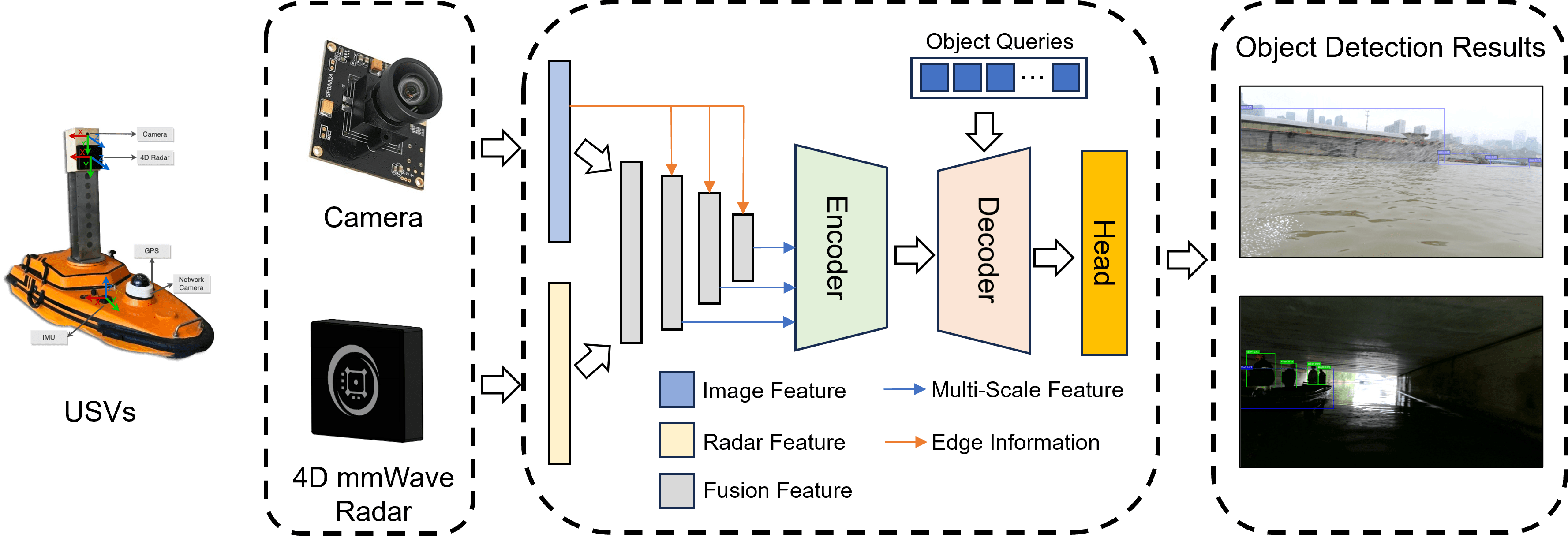}
    \caption{Overview of our USVs object detection method in complex water surface environments. Our method extracts and fuses features from the camera and 4D-mmWave radar, combines low-dimensional edge features, and fuses them with multi-scale, high-dimensional features to achieve robust water surface object detection under adverse conditions.}
    \label{Overview}
\end{figure}
\par Despite these advantages, existing USV detection methods have systematic limitations in environmental adaptability, multi-modal data fusion conflicts, and global feature extraction, which require comprehensive improvement. First, water surface environments introduce notable complexities in object detection due to blurred edges and significant scale variations. Current methods lack specialized mechanisms to enhance boundary perception and handle the diverse range of maritime objects, from small buoys to large vessels, resulting in reduced detection accuracy, especially in dynamic water conditions. Second, prevalent vision-radar fusion approaches suffer from cross-modal feature conflicts when directly integrating heterogeneous data. Additionally, 4D-mmWave radar systems generate sparse, irregular point clouds that are difficult to process effectively, as traditional convolutional neural networks (CNNs) struggle with these irregular spatial patterns, resulting in compromised feature extraction and representation. Third, existing water surface perception frameworks predominantly rely on CNN architectures optimized for local feature extraction, limiting their global context reasoning capabilities. The limited adoption of modern detection paradigms such as DEtection TRansformer (DETR) in maritime scenarios restricts the potential for enhanced contextual understanding and structural adaptability required for robust performance under diverse environmental conditions, including fog, rain, and extreme lighting variations~\cite{USVDETR}.

To address the problem of insufficient feature extraction and global modeling capability, which is necessary for robust multi-modal data fusion and adaptation to complex environments, we propose a robust DETR-based model WS-DETR that fuses vision and 4D-mmWave radar and is specifically designed for water surface environments. Fig.~\ref{Overview} illustrates the key design elements of our framework. Our method enhances edge information and aggregates multi-scale features. It also integrates adaptive fusion mechanisms to mitigate modal conflicts and generates refined object queries for accurate and robust water surface detection across various conditions. Our proposed method demonstrates the considerable model robustness on the WaterScenes dataset~\cite{Waterscenes}. The contributions of this study on methodologies and experiments are listed as follows:

\begin{itemize}
\begin{comment}
\end{comment}
\item To address blurred edges and scale variation challenges, we redesign the DETR backbone and encoder with two novel enhanced modules in the USV object detection task: a Multi-Scale Edge Information Integration (MSEII) module that enhances edge perception, and a Hierarchical Feature Aggregator (HiFA) that focuses and diffuses multi-scale contextual information, enabling accurate detection of objects with diverse scales.

\item  We propose an Adaptive Feature Interactive Fusion (AFIF) module to address the feature conflicts issue in multi-modal fusion. Additionally, we implement a radar backbone utilizing self-moving point representations for continuous convolution to process sparse and irregular point clouds from the 4D-mmWave radar, enabling efficient feature extraction.

\item We conduct extensive experiments on the vision-radar water surface object detection dataset WaterScenes~\cite{Waterscenes}, and the results demonstrate that our method achieves state-of-the-art (SOTA) performance across multiple metrics. Our method performs particularly well in adverse conditions, with ${\rm mAP_{50}}$ and ${\rm mAP_{50\text{-}95}}$ improving by 3.6\% and 2.5\% under adverse lighting and adverse weather conditions, respectively, compared to the current SOTA model.

\end{itemize}

\section{RELATED WORK}

\subsection{DETR Model}
%时态统一
DETR~\cite{carion2020end} introduces transformers into object detection by formulating it as a set prediction problem. It eliminates the need for anchor boxes and non-maximum suppression (NMS) through bipartite matching, enabling end-to-end training. However, DETR suffers from slow convergence, high computational cost, and difficulties in query optimization. 

Several variants have been proposed to address these issues. Deformable-DETR~\cite{zhu2021deformable} improves attention efficiency. DAB-DETR~\cite{liu2022dab} introduces 4D reference points, while DN-DETR~\cite{li2022dn} uses denoising strategies to accelerate training. DINO~\cite{dino} enhances detection performance by introducing contrastive denoising training and mixed query selection. RT-DETR~\cite{rt-detr} achieves real-time performance by optimizing the network structure for high-speed inference. Despite these advancements, DETR-based approaches have seen limited application in water surface scenarios, where challenges such as edge blurring and object scale variation remain unresolved.

\subsection{Detection Methods for USV}
Vision-based object detection methods have been widely deployed in unmanned surface vehicles (USVs)~\cite{vision}. However, due to the limitations of single-modal sensing, vision-radar fusion approaches have emerged as a research focal point to improve accuracy and robustness in maritime object detection~\cite{Waterscenes}.
Current methodologies exhibit notable performance trade-offs. ASY-VRNet~\cite{ASY-VRNET} emphasizes the processing speed at the cost of detection precision, while Mask-VRDet~\cite{Mask-VRDet} achieves higher precision but is impeded by excessive parameter complexity and computational demands. These approaches have not yet established an optimal balance between detection accuracy and cost, highlighting significant opportunities for our research.
Surface water environments introduce distinct challenges characterized by blurred edges, substantial variations in the scale of the object, and constantly changing conditions. Although existing research, such as BEMSNet~\cite{BEMSNet}, emphasizes the importance of sea surface object boundary detection. It primarily focuses on calm marine environments with large vessels while overlooking more complex scenarios. RISFNet~\cite{FLOW} addresses the challenge of small object detection, but its application is limited to single-category detection. Although Achelous~\cite{Achelous} is effective in river channel scenarios, it lacks robustness under adverse weather and lighting conditions commonly encountered on water surfaces. These limitations collectively underscore the need for a more comprehensive and adaptable detection framework tailored to the complexities of real-world water surface environments.

\section{METHODOLOGY}

\begin{figure*}[!t]
    \vspace{0.3cm}
    \centering
    \includegraphics[width=2\columnwidth]{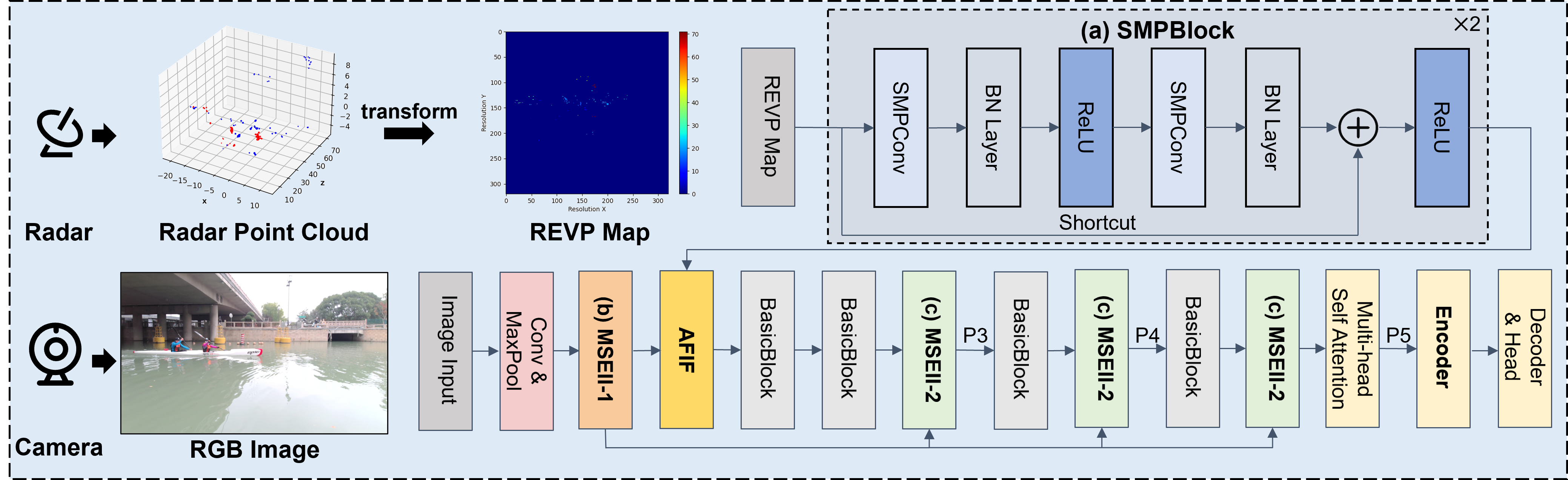}
    \caption{The network pipeline of WS-DETR. The model extracts multi-level features from both modalities, then performs fusion using the Adaptive Feature Interactive Fusion (AFIF) module, enriches high-dimensional features with low-level edge information via the Multi-Scale Edge Information Integration (MSEII) module, and finally applies Hierarchical Feature Aggregator (HiFA) module within the encoder for multi-scale feature fusion before detection.}
    \label{Framework}
\end{figure*}

\subsection{Overview}
WS-DETR is a novel object detection model that fuses vision and 4D-mmWave radar data based on the DETR structure. The complete network pipeline is summarized in Fig.~\ref{Framework}, showcasing the sequential stages from dual-branch input to detection output. The visual branch extracts preliminary features and edge information using convolutional layers and the Multi-Scale Edge Information Integration module, while the radar branch employs residual blocks with self-moving point representations (SMPBlock) to perform continuous convolution. The two modalities are fused via an Adaptive Feature Interactive Fusion module to produce enhanced image features. The enhanced image features are further refined by a ResNet-18~\cite{resnet} backbone with four stages of BasicBlocks. During this process, the Multi-Scale Edge Information Integration module injects low-level edge information into high-level multi-scale features. Then the multi-scale features are fused and diffused through the Hierarchical Feature Aggregator module inside the encoder. Finally, the fused features are fed into the decoder, and the detection head outputs the object category and bounding box.

\subsection{Pre-processing of Data and Dual-branch Backbone}
Since our task is based on camera-plane 2D detection, we transform the 3D radar point cloud onto the camera plane through a coordinate system. The point cloud generated by the 4D-mmWave radar has four features: range, elevation, velocity, and reflected power. Using these features, a radar data representation known as the REVP (range-elevation-velocity-power) map~\cite{ASY-VRNET} is employed, which is a four-channel image-like feature map.

Subsequently, the processed REVP map and image are fed into a dual-branch feature extraction network. The image first passes through the convolution layers and max pooling layer for preliminary feature extraction. Meanwhile, we use the self-moving point representation for continuous convolution (SMPConv)~\cite{smpconv} combined with residual connections for further improvement to extract irregular radar features.

SMPConv is a continuous convolution operator designed for irregular and sparse data, such as radar point clouds. Instead of using multi-layer perceptrons (MLP) to generate continuous kernels, SMPConv represents kernels via learnable moving points in continuous space. Each point has a set of learnable parameters $\phi$: trainable position $\mathbf{p}_i$, point feature $\mathbf{w}_i$, and radius $\mathbf{r}_i$. The kernel value at location $\mathbf{x}$ is computed by interpolating neighboring points as formulated in Eq.~(\ref{eq:smp}):
\begin{equation}
    SMP(\mathbf{x}; \phi) = \frac{1}{|\mathcal{N}(\mathbf{x})|} \sum_{i \in \mathcal{N}(\mathbf{x})} \left( 1 - \frac{\|\mathbf{x} - \mathbf{p}_i\|_1}{\mathbf{r}_i} \right) \mathbf{w}_i,
    \label{eq:smp}
\end{equation}
where $\mathcal{N}(\mathbf{x})$ denotes the set of neighboring points contributing to $\mathbf{x}$. The positions $\mathbf{p}_i$ are updated during training, allowing the receptive field to dynamically adapt to dense or sparse areas. Given the irregular nature of radar point clouds, SMPConv is particularly effective, as it can allocate more kernel points in dense regions and fewer in sparse regions, enabling efficient and adaptive convolution over non-grid data.

\subsection{Adaptive Feature Interactive Fusion (AFIF)} 
\begin{figure}[t]
    \vspace{0.3cm}
    \centering
    \includegraphics[width=\columnwidth]{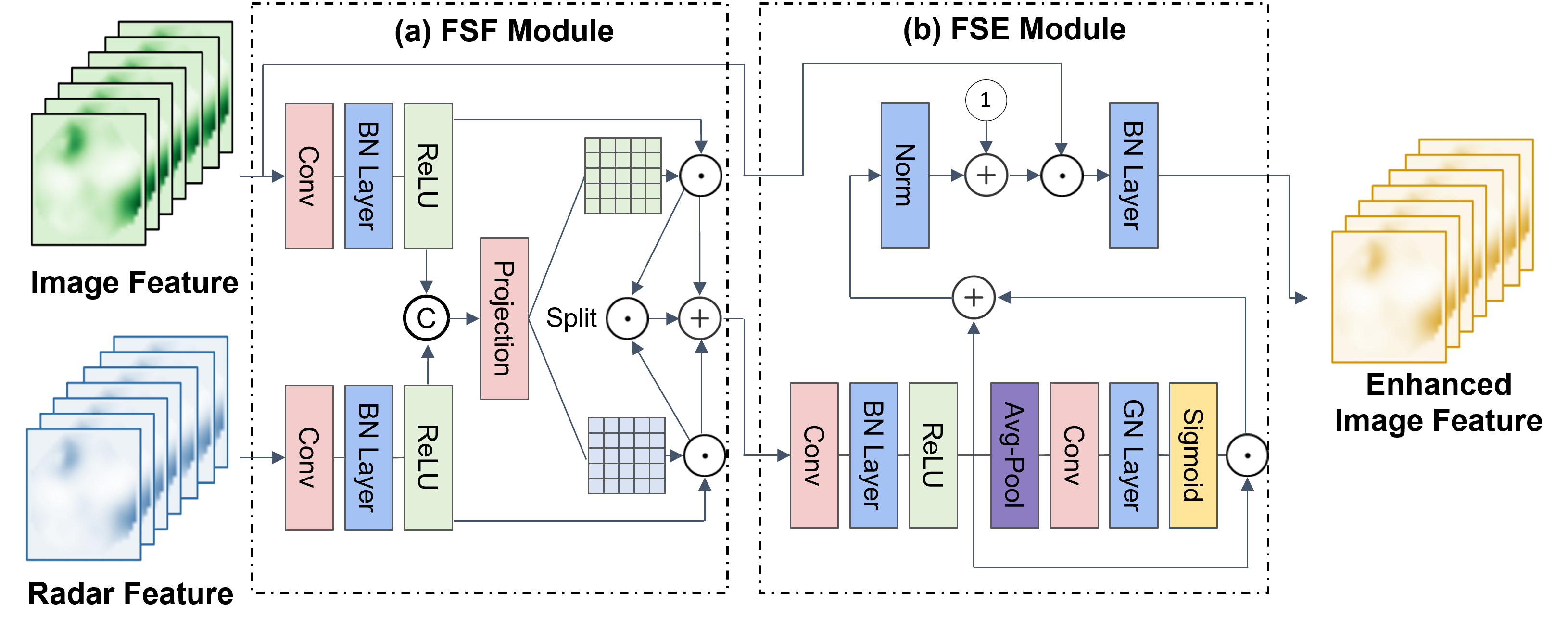}
    \caption{The structure of the Adaptive Feature Interactive Fusion (AFIF) module. The AFIF architecture comprises two cascaded stages: Feature Synchronization Fusion (FSF) for adaptive information supplementation of radar features and Feature Selection Enhancement (FSE) for cross-modal feature alignment enhancement of image features, which ultimately generates an enhanced image feature.}
    \label{AFIF}
\end{figure}

To address the challenge of multi-modal information fusion in collaborative object detection systems, an Adaptive Feature Interactive Fusion module is proposed to enable synergistic integration of 4D-mmWave radar point clouds with visual images. The Adaptive Feature Interactive Fusion module is a two-stage architecture. It first performs adaptive information supplementation on radar features through the Feature Synchronization Fusion module, which is illustrated in Fig.~\ref{AFIF}(a). Then it enhances image features using the selected radar features via the Feature Selection Enhancement module, which is illustrated in Fig.~\ref{AFIF}(b). This stage ultimately achieves effective cross-modal fusion.

In multi-modal feature fusion, low-entropy features may significantly degrade model performance during late training stages, not only weakening the complementarity of multi-modal features but also potentially introducing additional noise~\cite{FCMNet}. Consequently, effectively supplementing low-information features is critical for constructing robust multi-modal fusion methods. To address this challenge, we use a Feature Synchronization Fusion module for 4D-mmWave radar feature fusion and implementation, which initially establishes joint feature representation $\mathbf{F}_I$ and $\mathbf{F}_R$ through parallel convolutional pathways for feature extraction. Then a learnable fusion weight $\mathbf{W}_f$ is constructed by concatenating cross-modal features followed by cascaded convolutions, as defined in Eq.~(\ref{Wf}).
\begin{equation}
%W_{\text{f}} = \sigma(\text{Proj}(\text{concat}(F_I, F_R))) ,
\mathbf{W}_f = \sigma(\text{Proj}(\text{concat}(\mathbf{F}_I, \mathbf{F}_R))),
\label{Wf}
\end{equation}
where $\sigma$ denotes the sigmoid function, and \text{Proj} represents the projection layer that compresses the concatenated feature channels into two channels through cascaded convolutions. The final fused representation $\mathbf{F}_o$ captures both linear and nonlinear interactions through element-wise addition and multiplication between the feature maps and modality-specific attention weights. Specifically, these attention weights are derived from the two channels of the weight map $\mathbf{W}_f$, which are respectively applied to the radar and image features through element-wise multiplication to adaptively modulate their contributions. The final fused representation $\mathbf{F}_o$ is obtained via Eq.~(\ref{Fo}).

\begin{equation}
\begin{split}
\mathbf{F}_o = (\mathbf{W}_f[:, 0, :, :] \odot \mathbf{F}_I) + (\mathbf{W}_f[:, 1, :, :] \odot \mathbf{F}_R) \\ + (\mathbf{W}_f[:, 0, :, :] \odot \mathbf{F}_I \odot \mathbf{W}_f[:, 1, :, :] \odot \mathbf{F}_R) .
\end{split}
\label{Fo}
\end{equation}

The second stage performs spatial-aware feature enhancement through four sequential operations. First, the radar point cloud data $\mathbf{R}_p$ is transformed through $3 \times 3$ convolution into $\mathbf{R}'_p$, achieving channel dimension alignment and feature space mapping. Subsequently, we precisely calculate the channel attention weights $\boldsymbol{\omega}_A$ via global average pooling (GAP), convolution($\text{Conv}_{1\times1}$), group normalization (GN), and sigmoid function $\sigma$, as defined in Eq.~(\ref{wa}). 
\begin{equation}
%\omega_A = \sigma\left(\text{GN}\left(\text{Conv}_{1\times1}\left(\text{GAP}(R_p')\right)\right)\right).
\boldsymbol{\omega}_A = \sigma(\text{GN}(\text{Conv}_{1 \times 1}(\text{GAP}(\mathbf{R}'_p)))).
\label{wa}
\end{equation}

The attention weights $\boldsymbol{\omega}_A$ are applied to filter radar features through element-wise multiplication. In this process, unimportant channels in the radar features will be suppressed and important channels will be enhanced, thus achieving the adaptive selection of radar features, the filtered radar feature $\mathbf{R}_s$ is calculated as shown in Eq.~(\ref{Ratt}).
\begin{equation}
\mathbf{R}_s = \boldsymbol{\omega}_A \odot \mathbf{R}'_p.
\label{Ratt}
\end{equation}

Then, the normalized output combines residual radar information to enhance image features $\mathbf{I}$. The enhanced image features $\mathbf{I}_e$ are computed as described in Eq.~(\ref{Ie}). 
\begin{equation}
\mathbf{I}_e = (1 + \text{Norm}(\mathbf{R}_s + \mathbf{R}'_p)) \odot \mathbf{I}.
\label{Ie}
\end{equation}

AFIF is a two-stage fusion module designed for vision–radar multi-modal object detection. The first stage captures modality-specific representations and models cross-modal interactions through linear and nonlinear interactions. The second stage further aligns the features via spatial-aware enhancement, leveraging radar geometry to enhance visual features. Unlike conventional implicit fusion strategies, AFIF explicitly decouples low-level geometric alignment and high-level semantic integration, effectively reducing conflicts in cross-modal feature fusion.

\subsection{Multi-Scale Edge Information Integration (MSEII)}
 
In water surface environments, dynamic environments often cause edge blurring in images. Traditional object detection networks lack explicit mechanisms to focus on edge information, while object localization is highly dependent on the integrity of edge features. In existing network architectures, the fine edge information contained in shallow features struggles to effectively propagate to deeper layers, and the multi-scale feature fusion process tends to dilute edge feature representation. Meanwhile, directly extracting edge information from raw images and passing it through the backbone would introduce excessive noise due to irrelevant background details. Since shallow convolutional layers naturally filter out unnecessary background information, the Multi-Scale Edge Information Integration module extracts edge features in the shallow layer, as illustrated in Fig.~\ref{Framework}(b).

To extract edge features, we employ Sobel convolution, which computes the horizontal gradient $\mathbf{G}_x$ and the vertical gradient $\mathbf{G}_y$ of the image:
\begin{equation}
    \mathbf{G}_x = \mathbf{I} * \mathbf{K}_x, \quad \mathbf{G}_y = \mathbf{I} * \mathbf{K}_y,
\end{equation}
\begin{comment}
where the Sobel kernels are given by Eq.~(\ref{kx}):
\begin{equation}
    \mathbf{K}_x = \begin{bmatrix}1&2&1\\0&0&0\\-1&-2&-1\end{bmatrix},\quad 
    \mathbf{K}_y = \begin{bmatrix}1&0&-1\\2&0&-2\\1&0&-1\end{bmatrix},   
    \label{kx}
\end{equation}
\end{comment}
after computing the magnitude of the gradient $\mathbf{G}=\sqrt{\mathbf{G}_x^2 + \mathbf{G}_y^2}$, given that different objects and environmental textures exist on varying scales, a single-scale edge map is insufficient. Thus, we generate multi-scale edge maps by progressively downsampling the extracted edge features, as described in Eq.~(\ref{Gl}).
\begin{equation}
    \mathbf{G}^{(l)} = \text{MaxPool}(\mathbf{G}^{(l-1)}), \quad l \in \{1, 2, \dots, L\},
\label{Gl}
\end{equation}
where $l$ denotes the edge feature scale, and max pooling is used instead of average pooling due to its superior ability to preserve strong local edges.

Multi-Scale Edge Information Integration module further integrates edge-aware features with backbone features, which is shown in Fig.~\ref{Framework}(c). Specifically, a learnable convolution combines semantic and edge information, followed by $3 \times 3$ convolutions for local refinement, and a $1 \times 1$ convolution for dimensionality reduction while preserving key details. Finally, the output is added to the output of the BasicBlock to prevent information loss.

Based on the above operations, the Multi-Scale Edge Information Integration module enhances object detection by effectively integrating edge-aware features into the backbone network. 
\subsection{Encoder with  Hierarchical Feature Aggregator (HiFA) }
\begin{comment}
 \begin{figure}[h]
    \centering
    \includegraphics[width=\columnwidth]{HEAF.png}
    \caption{Hierarchical Feature Aggregator (HiFA) Module.}
    \label{HiFA}
\end{figure}   
\end{comment}

\begin{figure}[t]
    \vspace{0.3cm}
    \centering
    \includegraphics[width=\columnwidth]{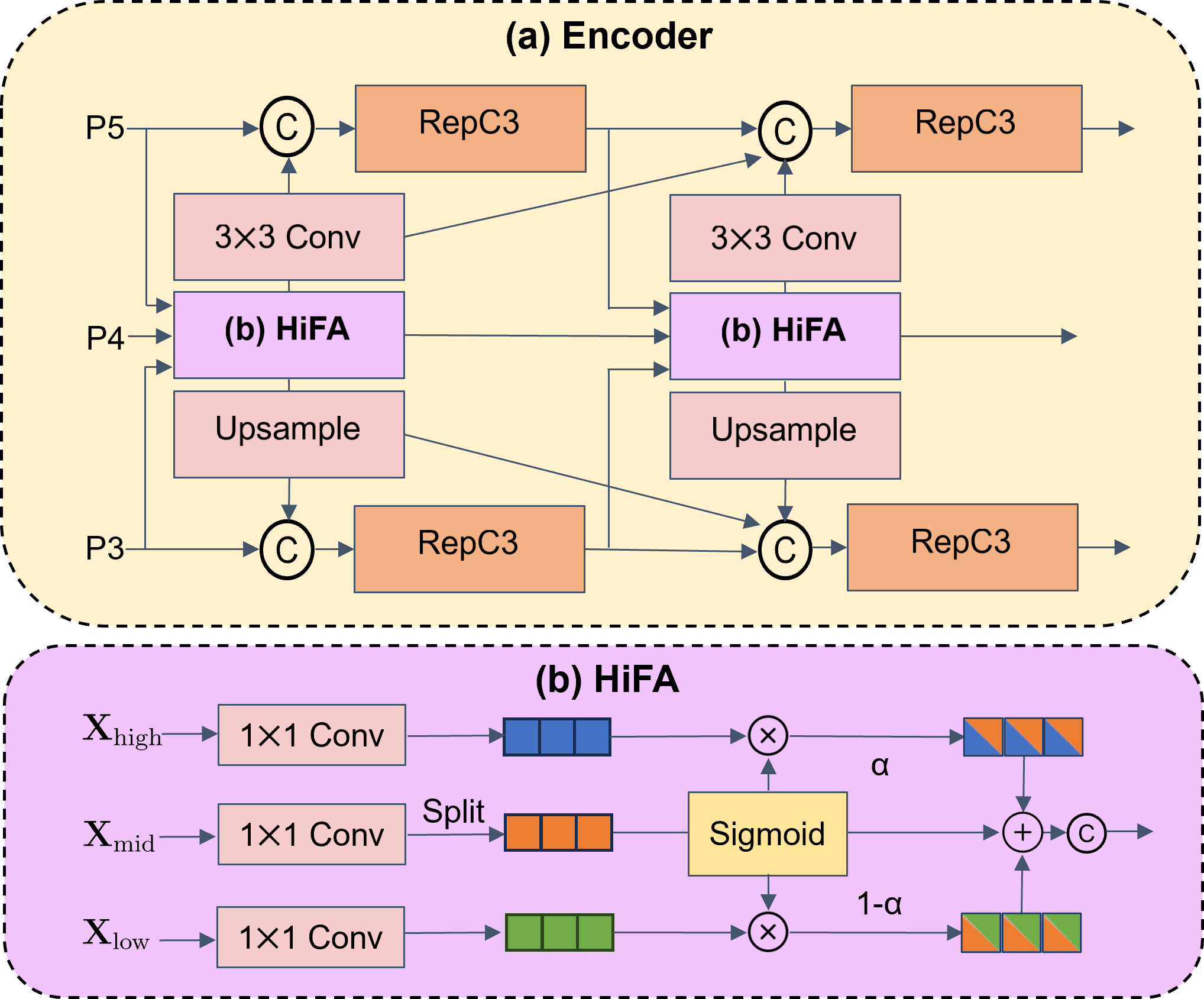}
    \caption{The structure of the Hierarchical Feature Aggregator (HiFA) module and its integration in the encoder. HiFA fuses multi-scale features via scale-specific transformations and a dynamic weighting mechanism. The encoder employs HiFA modules to enhance cross-scale information flow, followed by upsampling and convolution to diffuse features across levels, enabling robust object detection under scale variations.}
    \label{Encoder}
\end{figure}

Since USV object detection scenes have the characteristics of large changes in object scale, we design a new DETR encoder structure to enhance cross-scale information flow, which is illustrated in Fig.~\ref{Encoder}(a).

The multi-scale features (P3, P4, P5) coming from the backbone network, as shown in Fig.~\ref{Framework}, first go through the Hierarchical Feature Aggregator module for feature interaction. Then upsample and conv layer are utilized for information transfer between different scales. Subsequently, the features of each scale will enter the RepC3~\cite{rt-detr} module for further feature enhancement and fusion with the outputs of other scales. Through this multi-level top-down and bottom-up connection and enhancement mechanism, the encoder can fully exchange information between feature maps of different resolutions and obtain a richer and more hierarchical feature representation~\cite{HCFNET}.

For the Hierarchical Feature Aggregator module, which is shown as Fig.~\ref{Encoder}(b), and given a multi-scale input $\{\mathbf{X}_{\text{low}}, \mathbf{X}_{\text{mid}}, \mathbf{X}_{\text{high}}\}$, the module first applies a convolutional layer that transforms features at three different scales to a uniform size.
\begin{comment}
  \begin{equation}
\begin{cases}
\mathbf{X}'_{\text{low}} = \text{Conv}_{1 \times 1}(\mathbf{X}_{\text{low}}) \uparrow_{\text{bi}}, \\
\mathbf{X}'_{\text{high}} = \text{Conv}_{1 \times 1}(\mathbf{X}_{\text{high}}) \downarrow_{\text{conv3}}, \\
\mathbf{X}'_{\text{mid}} = \text{Conv}_{1 \times 1}(\mathbf{X}_{\text{mid}}),
\label{trans}
\end{cases}
\end{equation}  
where $\uparrow_{\text{bi}}$ represents bilinear upsampling, and $\downarrow_{\text{conv3}}$ represents 3×3 convolution.
\end{comment}

The transformed features $\{\mathbf{X}'_{\text{low}},\mathbf{X}'_{\text{high}}, \mathbf{X}'_{\text{mid}}\}$ are partitioned into $K$ chunks ($K=3$) for parallel processing, each branch achieves adaptive feature aggregation through weighted calculation, which is performed according to Eq.~(\ref{fused}):

\begin{equation}
\mathbf{X}_{\text{fused}}^{(k)} = \sigma(\mathbf{X}_{\text{mid}}'^{(k)}) \odot \mathbf{X}_{\text{high}}'^{(k)} + [1-\sigma(\mathbf{X}_{\text{mid}}'^{(k)})] \odot \mathbf{X}_{\text{low}}'^{(k)},
\label{fused}
\end{equation}
where $\sigma$ is sigmoid function, $\sigma(\mathbf{X}_{\text{mid}}'^{(k)})$ denotes the dynamic attention weight $\alpha$. The fused features are concatenated and processed, which is helpful for the model to recognize features at different scales.

\subsection{Decoder and Head}

Inspired by the RT-DETR~\cite{rt-detr}, our model uses a 3-layer transformer decoder structure. The high-quality initial queries are selected through an uncertainty-minimizing approach. The decoder then optimizes object predictions with multi-head attention mechanisms. We introduce a feature uncertainty metric $U(\hat{\mathbf{X}})$ for better optimization, $\hat{\mathbf{X}}$ denotes the encoder feature, which is computed based on Eq.~(\ref{U}). This metric measures the difference between localization $P(\hat{\mathbf{X}})$ and classification $C(\hat{\mathbf{X}})$ prediction distributions. The loss function combines box regression and classification components, as defined in Eq.~(\ref{L}). Then, the auxiliary losses from multiple decoder layers are utilized to accelerate training convergence. This design achieves both efficiency and accuracy in object detection.

\begin{equation}
U(\hat{\mathbf{X}}) = \left\|P(\hat{\mathbf{X}}) - C(\hat{\mathbf{X}})\right\|,
    \label{U}
\end{equation}
\begin{equation}
L(\hat{\mathbf{X}}, \hat{\mathbf{Y}}, \mathbf{Y}) =
L_{\text{box}}(\hat{\mathbf{b}}, \mathbf{b}) +
L_{\text{cls}}(U(\hat{\mathbf{X}}), \hat{\mathbf{c}}, \mathbf{c}),
    \label{L}
\end{equation}
where $L_{\text{box}}$ is composed of $\ell_1$ and GIoU loss for bounding box regression, while $L_{\text{cls}}$ is the standard cross-entropy loss for classification. $\hat{\mathbf{Y}}$ and $\mathbf{Y}$ denote the predicted and ground truth labels, respectively, and $\hat{\mathbf{c}}$ and $\hat{\mathbf{b}}$ represent the predicted category and bounding box.

\section{EXPERIMENTS}

\subsection{Experimental Setup}

\textbf{Datasets.} We train WS-DETR on the WaterScenes dataset, which includes 54,120 groups of RGB images and single frame radar point clouds in various water surface environments, and contains more than 200,000 objects. All images are in 1920 × 1080 resolution and contain various maritime objects, including piers, buoys, sailors, ships, boats, vessels, and kayaks. Following the guidelines outlined by the WaterScenes dataset, we divide it into training, validation, and test sets with a ratio of 7:2:1. For the test set, we specifically select samples under adverse lighting and weather conditions, based on varying lighting and weather scenarios, to evaluate the accuracy and robustness of the model under challenging conditions.

\textbf{Implementation Details.} The images and REVP maps are resized to 640 $\times$ 640 pixels. WS-DETR is trained from scratch for 150 epochs with a batch size of 8 and an initial learning rate of 0.0001, using the AdamW optimizer on an RTX 4090 GPU.

\textbf{Evaluation Metrics.} The mean Average Precision (mAP) is used as the primary evaluation metric, with both ${\rm mAP_{50}}$ and ${\rm mAP_{50\text{-}95}}$ adopted. ${\rm mAP_{50}}$ is computed at an IoU threshold of 0.5, reflecting performance under loose localization, while ${\rm mAP_{50\text{-}95}}$ averages precision over IoU thresholds from 0.5 to 0.95 in steps of 0.05, providing a more rigorous evaluation.
\begin{comment}
\begin{equation}
{\rm mAP_{50}}= \frac{1}{N} \sum_{i=1}^{N} \text{AP}^{i}_{50},
\label{map50}
\end{equation}
\begin{equation}
{\rm mAP_{50\text{-}95}} = \frac{1}{10} \sum_{t=0.5}^{0.95} {\rm mAP_{t}},
\label{map5095}
\end{equation}
where $N$ denotes the total number of object classes.
\end{comment}

\begin{table*}[t]
\vspace{0.3cm}
\centering
\resizebox{\textwidth}{!}{
\begin{threeparttable}
\caption{Comparison of WS-DETR with other object detection models on the WaterScenes dataset using ${\rm mAP_{50}}$ and ${\rm mAP_{50\text{-}95}}$ metrics. In the modalities column, C denotes camera input, and R denotes 4D-mmWave radar input. Per-class detection accuracy is presented using ${\rm mAP_{50}}$.}
\label{tab:wsdetr-comparison}
\begin{tabular}{cc|cc|ccccccc|cc}
\hline
\textbf{Model} & \textbf{Modalities} & \textbf{Params(M)} & \textbf{FLOPS(G)} & \textbf{Pier} & \textbf{Buoy} & \textbf{Sailor} & \textbf{Ship} & \textbf{Boat} & \textbf{Vessel} & \textbf{Kayak} & $\textbf{mAP}_{50\mbox{-}95}$ & $\textbf{mAP}_{50}$ \\ \hline

Faster R-CNN~\cite{fasterrcnn}\tnote{1} & C & 42.0 & 180.0 & 81.3 & 78.4 & 75.6 & 93.0 & 88.9 & 92.2 & 58.2 & 47.8 & 81.1 \\ 
CenterNet~\cite{centernet}\tnote{1}  & C & 32.3 & 207.0 & 83.0 & 80.1 & 79.3 & 92.7 & 89.5 & 93.1 & 62.9 & 54.7 & 82.9 \\ 
\hline
DAB-DETR~\cite{liu2022dab} & C & 43.7 & 33.9 & 68.7 & 55.0 & 53.7 & 89.7 & 84.5 & 90.7 & 67.5 & 38.8 & 72.8 \\ 
\makecell{DAB-Deformable-\\DETR~\cite{liu2022dab}}  & C & 47.8 & 114.7 & 60.6 & 56.0 & 50.9 & 70.6 & 74.0 & 81.1 & 70.8 & 47.3 & 83.9 \\ 
Deformable DETR~\cite{zhu2021deformable}\tnote{1} & C & 40.0 & 173.0 & 83.9 & 82.2 & 80.2 & 92.9 & 89.4 & 92.7 & 66.8 & 56.5 & 84.0 \\ 
RT-DETR~\cite{rt-detr} & C & 20.0 & 57.0 &92.9&88.2&87.3&97.3&92.3&97.5&66.4& 59.8 & 88.9\\ 
DINO~\cite{dino} & C & 52.3& 47.6 & 89.9 & 84.0 & 80.9 & 95.1 & 91.3 & 95.8 & 83.8 & 57.8 & 88.7 \\ 
\hline
YOLOX-M~\cite{yolox}\tnote{1}& C & 25.3 & 73.8 & 85.1 & 81.1 & 80.5 & 91.4 & 89.5 & 92.1 & 76.1 & 57.8 & 85.1\\ 
YOLOv8-M~\cite{yolov8}\tnote{1}& C & 25.9 & 79.3 & 80.6 & 84.3 & 82.1 & 93.7 & 90.8 & 95.8 & 62.5 & 59.2 & 84.4  \\  
YOLOv9-M~\cite{yolov9}& C & 20.2 & 77.9 &88.0&86.3&84.6&94.7&93.5&96.1&66.2& 61.6 & 87.1  \\ %epoch=50
YOLOv10-M~\cite{yolov10} & C & 16.5 & 64.0 &88.9&87.7&87.4&95.6&93.7&97.0&67.4& 62.9 & 88.2  \\%epoch=40
YOLOv11-M~\cite{yolov11} & C & 20.1 & 67.7 &90.0&88.4&87.2&96.2&\textbf{94.3}&97.1&68.1& 63.9 & 88.9  \\%epoch=40
%epoch=40
\hline
YOLOX-M~\cite{Waterscenes}\tnote{1} & C + R & - & - & 85.5 & 82.2 & 81.3 & 92.9 & 91.3 & 92.5 & 77.1 & 59.5 & 86.1  \\
%YOLOX-M [] & C + R3 & 60.3 & - & - & 87.1 & 47.8 & 81.5 & 83.5 \\
YOLOv8-M~\cite{Waterscenes}\tnote{1} & C + R & - & - & 86.2 & 85.9 & 85.1 & 94.6 & 91.2 & 95.0 & \textbf{77.9}& 61.2 & 88.0 \\
%YOLOv8-M [] & C + R3 & 62.5 & - & - & 84.5 & 47.8 & 82.1 & 84.2 \\
Achelous~\cite{Achelous}\tnote{2} & C + R & 7.1 & 14.6 &-&-&-&-&-&-&-& 40.5 & 70.8  \\
%ASY-VRNet~\cite{ASY-VRNET} & C + R & 29.0 & 40.5 &88.2&70.6&63.3&88.2&81.6&92.2&6.9& 38.0 & 68.9  \\
ASY-VRNet~\cite{ASY-VRNET}\tnote{3} & C + R & 4.1 & 3.26 &-&-&-&-&-&-&-&42.8&-  \\
\hline
Ours & C + R & 22.6 & 79.8 &\textbf{94.1}&\textbf{89.8}&\textbf{92.9}&\textbf{97.8}&92.3&\textbf{98.3}&74.7& \textbf{64.5} & \textbf{91.4} \\
\hline
\end{tabular}
\begin{tablenotes}
\footnotesize
\item[1] The results of these models are adopted from the paper~\cite{Waterscenes}.
\item[2] The results of Achelous is adopted from the paper~\cite{Achelous}.
\item[3] The results of ASY-VRNet is adopted from the paper~\cite{ASY-VRNET}.
\end{tablenotes}
\end{threeparttable}
}
\end{table*}

\begin{figure}[t]
    \centering
    \includegraphics[width=\columnwidth]{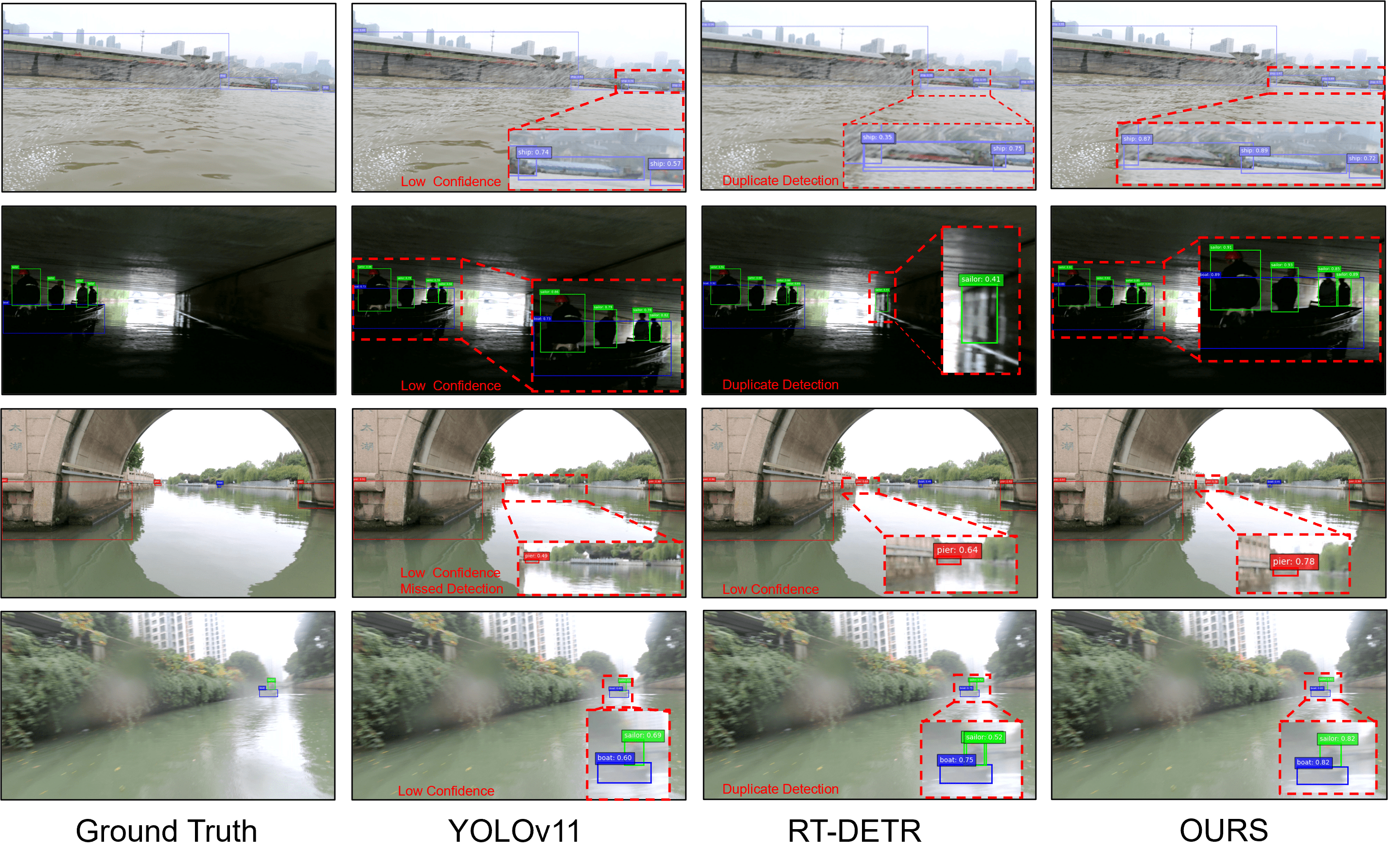}
    \caption{Comparison of experimental results. We select YOLOv11 and RT-DETR, which achieve strong performance on the dataset, to compare with our proposed model WS-DETR under different environmental conditions. The first row shows results in rainy scenes, the second in low-light environments, the third in scenes with multi-scale objects, and the last in scenes with edge-blurred objects.
}
    \label{visual}
\end{figure}

\subsection{Comparison of WS-DETR with Other Models}

To comprehensively evaluate the performance of WS-DETR, we compare it against four categories of mainstream object detection models: traditional CNN-based architectures (Faster R-CNN~\cite{fasterrcnn}, CenterNet~\cite{centernet}), DETR-based models (Deformable DETR~\cite{centernet}, RT-DETR~\cite{rt-detr}, DINO~\cite{dino}, DAB-DETR~\cite{liu2022dab}, DAB-Deformable-DETR~\cite{liu2022dab}), YOLO series models (YOLOX~\cite{yolox}, YOLOv8~\cite{yolov8}, YOLOv9~\cite{yolov9}, YOLOv10~\cite{yolov10}, YOLOv11~\cite{yolov11}), and vision-radar fusion-based models (Achelous~\cite{Achelous}, ASY-Net~\cite{ASY-VRNET}, improved YOLOv8, improved YOLOX). Among them, Faster R-CNN, CenterNet, DAB-DETR, DAB-Deformable-DETR, Deformable DETR, and DINO adopt ResNet-50 as the backbone, RT-DETR uses ResNet-18, and Achelous adopts the Achelous-EF-GDF-PN-S2 variant. All YOLO series models use medium-size (M) configurations. The ${\rm mAP_{50}}$ and ${\rm mAP_{50\text{-}95}}$ are used as evaluation metrics, with the experimental results shown in Table~\ref{tab:wsdetr-comparison}.

Our proposed model demonstrates significant advantages in terms of accuracy and computational efficiency. Table~\ref{tab:wsdetr-comparison} shows that our model achieves 91.4\% ${\rm mAP_{50}}$ and 64.5\% ${\rm mAP_{50\text{-}95}}$, outperforming all comparison models. Notably, WS-DETR maintains this high accuracy while using only 22.6M parameters and 79.8G FLOPS, which is substantially lower than traditional CNN models like Faster R-CNN (42.0M) and better than most DETR-based models, such as DAB-Deformable-DETR (47.8M). Looking at category-specific detection performance, WS-DETR achieves high accuracy across all seven categories, with particularly best performance in the Pier (94.1\%), Buoy (89.8\%), Sailor (92.9\%), Ship (97.8\%), and Vessel (98.3\%) categories. These results demonstrate that WS-DETR successfully achieves higher detection accuracy without significantly increasing the parameter count and computational complexity, reflecting an excellent balance between efficiency and performance in the model design, especially compared to other vision-radar fusion models.

\subsection{Robustness Experiment}
WS-DETR also demonstrates remarkable performance advantages under adverse lighting and weather conditions. As illustrated by Table~\ref{2}, our model achieves 85.7\% ${\rm mAP_{50}}$ under adverse lighting and 84.9\% ${\rm mAP_{50}}$ under adverse weather conditions, outperforming all comparison models. For instance, WS-DETR outperforms the best-performing model YOLOv10-M by 3.6\% and 2.5\% in ${\rm mAP_{50}}$, demonstrating the effectiveness of our method in addressing complex environmental interferences. These results indicate that WS-DETR not only excels under normal conditions but also maintains high-precision detection capabilities in challenging scenarios such as insufficient lighting and rainy environments, providing a more reliable solution for practical maritime applications.

Fig.~\ref{visual} presents a visual comparison of top-performing models on the WaterScenes dataset. As shown in the first two rows, our method accurately completes object detection tasks even under rain and dark lighting conditions, avoiding the multiple detections exhibited by RT-DETR and the missed detections seen with YOLOV11-M. The last two rows demonstrate that when facing objects with significant scale variations and blurred boundaries, our method successfully detects all objects with the highest confidence scores, further validating the effectiveness of our approach.

\begin{table}[!ht]
\centering
\caption{Evaluation of models under adverse lighting and weather conditions using ${\rm mAP_{50}}$.}
\begin{tabular}{cc|cc}
\hline
\textbf{Model} & \textbf{Modalities} &\textbf{\makecell[c]{Adverse\\lighting}} & \textbf{\makecell[c]{Adverse\\weather}} \\ \hline

Faster R-CNN & C & 69.4 & 71.1 \\ 
CenterNet  & C & 72.2 & 73.7 \\ 
\hline
Deformable DETR & C & 74.5 & 76.2 \\ 
RT-DETR & C & 82.5 & 80.1 \\ 
DINO & C & 79.1 & 81.9 \\ 
%DINO [] & C & - & - \\ 
DAB-DETR & C & 60.7 & 60.3 \\ 
\makecell{DAB-Deformable-\\DETR} & C & 69.4 & 71.8 \\ 
\hline
YOLOX-M & C & 77.4 & 78.9 \\ 
YOLOv8-M & C & 74.8 & 79.5 \\  
YOLOv9-M & C & 83.3 & 78.8 \\
YOLOv10-M & C & 82.1 & 82.4 \\
YOLOv11-M & C & 82.3 & 81.8 \\
\hline
YOLOX-M & C + R & 79.8 & 82.5 \\
%YOLOX-M & C + R3 & 60.3 & - & - & 87.1 & 47.8 & 81.5 & 83.5 \\
YOLOv8-M & C + R & 80.1 & 82.4 \\
%YOLOv8-M [] & C + R3 & 62.5 & - & - & 84.5 & 47.8 & 82.1 & 84.2 \\
%Achelous & C + R & - & - \\
%ASY-VRNet & C + R & - & - \\
\hline
Ours & C + R & \textbf{85.7} & \textbf{84.9} \\
\hline
\label{2}
\end{tabular}
\end{table}

\begin{table}[!ht]
    \centering
    \caption{Ablation Experiments of OUR Method}
    \begin{tabular}{c l l l l l}
    \hline
        \textbf{Method} & \textbf{AFIF} & \textbf{MSEII} & \textbf{HiFA} &$\textbf{mAP}_{50\mbox{-}95}$ & $\textbf{mAP}_{50}$ \\ 
        \multirow{8}{*}{WS-DETR}
        & - & - & - & 59.8 & 88.9 \\
        & \checkmark & - & - & 60.9 & 89.7 \\
        & - & \checkmark & - & 60.8 & 89.5 \\
        & - & - & \checkmark & 60.4 & 89.3 \\
        & \checkmark & \checkmark & - & 62.2 & \textcolor{blue}{90.8} \\
        & \checkmark & - & \checkmark & \textcolor{blue}{62.5} & 90.7 \\    
        & - & \checkmark & \checkmark & 61.4 & 90.5 \\   
        & \checkmark & \checkmark & \checkmark & \textcolor{red}{64.5} & \textcolor{red}{91.4} \\   
        
        \hline
    \end{tabular}
    \label{ablation}
\end{table}

\begin{table}[!ht]
    \centering
    \caption{COMPARISON WITH OTHER Radar BACKBONE DESIGN}
    \begin{tabular}{l l l l l}
    \hline
        \textbf{Backbone} & \textbf{Params(M)} &$\textbf{mAP}_{50\mbox{-}95}$ & $\textbf{mAP}_{50}$ \\ \hline
        ResNet-18 & 22.7 & 63.1 & 90.8  \\ 
        Ours & 22.6 & \textbf{64.5} & \textbf{91.4} \\ \hline
    \end{tabular}
    \label{radar}
\end{table}

\subsection{Ablation Experiments}

To validate the effectiveness of our proposed components, we conduct comprehensive ablation experiments as detailed in Table~\ref{ablation}. RT-DETR is selected as the baseline model with ResNet-18 backbone, which achieves 59.8\% ${\rm mAP_{50\text{-}95}}$ and 88.9\% ${\rm mAP_{50}}$. Among individual components, AFIF delivers the most substantial improvement. When integrating these components, we observe pronounced synergistic effects, with the dual combination of AFIF and HiFA exhibiting superior performance among two-component configurations. Notably, our full model incorporating all three components significantly outperforms all other variants, achieving 64.5\% ${\rm mAP_{50\text{-}95}}$ and 91.4\% ${\rm mAP_{50}}$, representing substantial improvements of 4.7\% and 2.5\% over the baseline, respectively. These results convincingly demonstrate that each proposed component contributes meaningfully to detection performance, with their synergistic combination yielding optimal accuracy.

Additionally, Table~\ref{radar} compares our radar backbone network composed of SMPConv and residual structure with ResNet-18 BasicBlock backbone. The results demonstrate that compared to the conventional ResNet-18 BasicBlock backbone, our method achieves accuracy improvements while maintaining a similar number of parameters, highlighting our model's efficiency-performance balance.

\begin{figure}[t]
    \vspace{0.3cm}
    \centering
    \includegraphics[width=\linewidth]{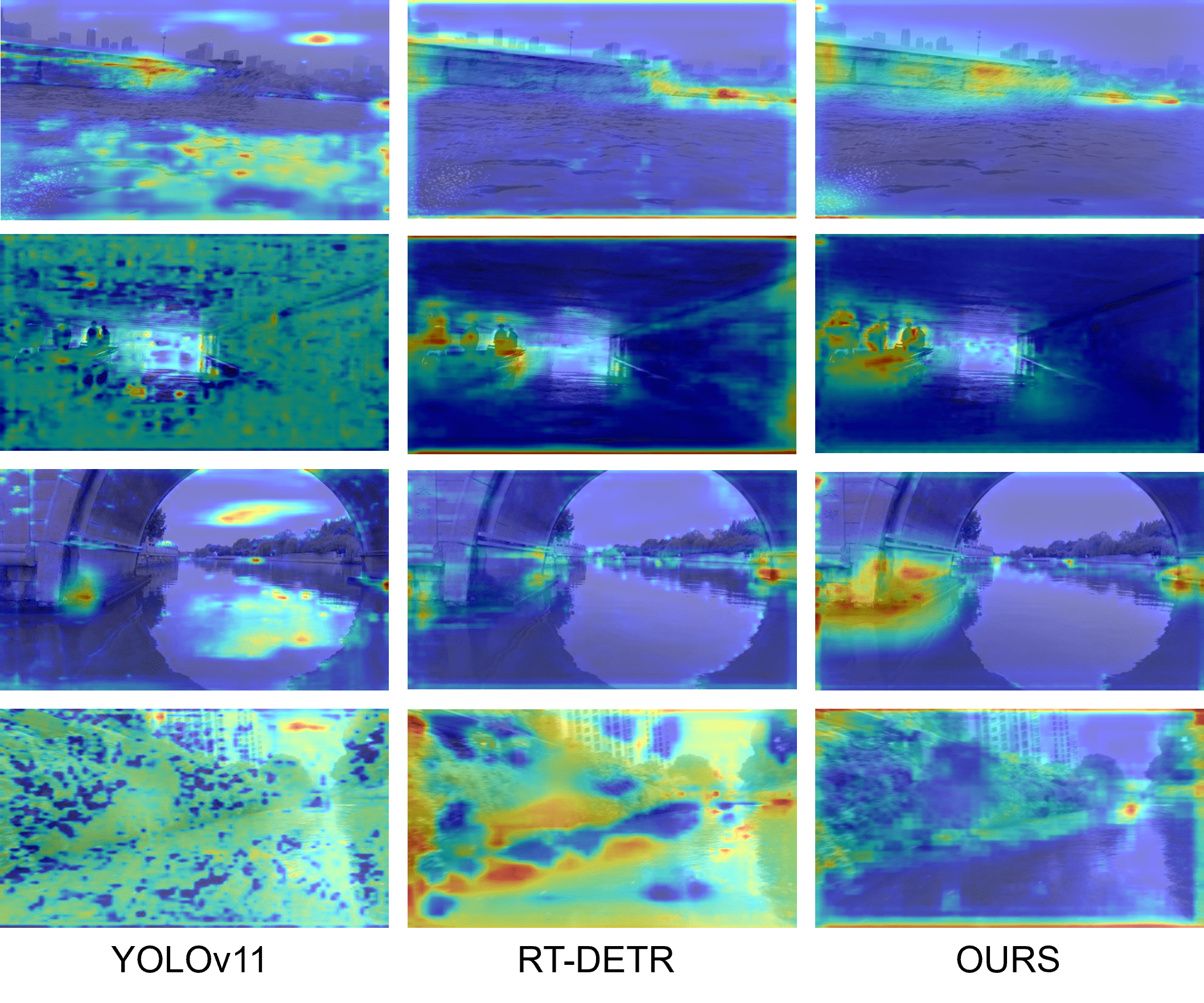}
    \caption{Class-specific heatmaps visualized using GradCAM++~\cite{Grad-CAM++} on the decoder input layers of each model. These heatmaps highlight how multi-scale features are aggregated for object detection.}
    \label{heatmap}
\end{figure}

\subsection{Visualization analysis}

To further investigate the effectiveness of our method, we apply GradCAM++~\cite{Grad-CAM++} to the decoder input layers of YOLOv11-M, RT-DETR, and our proposed model. As shown in Fig.~\ref{heatmap}, WS-DETR consistently produces more focused and complete attention regions across different challenging environments, including rainy weather, low-light conditions, multi-scale objects, and edge-blurred scenes.

In the first and second rows (rainy and dim conditions), WS-DETR exhibits clear and accurate attention to true object areas, while other models show dispersed or background-biased responses. In the third row, facing scale variation, our model better captures both small and large objects. In the last row, where edge information is weak, our model still maintains stable and complete object localization. These results demonstrate the robustness and accuracy of our method.

\begin{comment} 
   \begin{figure}[thpb]
      \centering
      \framebox{\parbox{3in}{We suggest that you use a text box to insert a graphic (which is ideally a 300 dpi TIFF or EPS file, with all fonts embedded) because, in an document, this method is somewhat more stable than directly inserting a picture.
}}
      \includegraphics[scale=0.3]{overview.png}
      \caption{Inductance of oscillation winding on amorphous
       magnetic core versus DC bias magnetic field}
      \label{figurelabel}
   \end{figure}
\end{comment}

\section{CONCLUSIONS}
In this paper, we propose a robust vision–radar fusion object detection model WS-DETR designed for complex water surface environments. By introducing a fusion strategy with the Adaptive Feature Interactive Fusion (AFIF) module, edge-aware enhancement via the Multi-Scale Edge Information Integration (MSEII) module, and hierarchical multi-scale aggregation through the Hierarchical Feature Aggregator (HiFA) module, our model effectively addresses challenges such as cross-modal conflicts, blurred boundaries, and scale variation. Extensive experiments on the WaterScenes dataset demonstrate that WS-DETR achieves SOTA performance under both normal and adverse conditions while maintaining strong generalization and efficiency.

In future work, we plan to explore the real-time deployment of WS-DETR in practical USV systems and further optimize the architecture toward a more lightweight design. These efforts aim to enhance the applicability of our approach in resource-constrained and dynamic aquatic environments.


%%%%%%%%%%%%%%%%%%%%%%%%%%%%%%%%%%%%%%%%%%%%%%%%%%%%%%%%%%%%%%%%%%%%%%%%%%%%%%%%

%%%%%%%%%%%%%%%%%%%%%%%%%%%%%%%%%%%%%%%%%%%%%%%%%%%%%%%%%%%%%%%%%%%%%%%%%%%%%%%%

%%%%%%%%%%%%%%%%%%%%%%%%%%%%%%%%%%%%%%%%%%%%%%%%%%%%%%%%%%%%%%%%%%%%%%%%%%%%%%%%
%\section*{APPENDIX}

\section*{ACKNOWLEDGMENT}
This work was supported by the National Natural Science Foundation of China under Grant No. 62433014. The authors would like to thank TÜV SÜD for the kind and generous support. We are also grateful for the efforts from our colleagues in Sino German Center of Intelligent Systems. We also express our gratitude for the assistance provided by Mr. Yujian Mo.

%%%%%%%%%%%%%%%%%%%%%%%%%%%%%%%%%%%%%%%%%%%%%%%%%%%%%%%%%%%%%%%%%%%%%%%%%%%%%%%%

\bibliography{root}

\bibliographystyle{unsrt}
\end{document}